\documentclass{article}
\usepackage[utf8]{inputenc}
\usepackage{todonotes}
\usepackage[colorlinks]{hyperref}
\usepackage{url}
\usepackage{scrextend}
\usepackage{rotating}
\usepackage{subcaption}

\title{Monash University, UEA, UCR\\ Time Series Extrinsic Regression Archive}
\author{Chang Wei Tan \and Christoph Bergmeir \and Fran\c{c}ois Petitjean \and Geoffrey I. Webb}
\date{Faculty of Information Technology, Monash University}

\newcommand{\arima}{\textsc{ARIMA}}

\newcommand{\etal}{\textit{et al.}}

\graphicspath{figures}

\begin{document}
\maketitle


\section{Introduction}
Machine Learning research generally relies on a good benchmarking archive. 
The well-known, publicly available machine learning dataset repository from the University of California Irvine (UCI) contains more than 450 datasets from various domains and areas \cite{Dua:2019}.
This repository has benefited the development of many state-of-the-art machine learning algorithms.
Similar phenomena have also been observed for time series research.
Time series research has gathered lots of interests in the last decade, 
especially for \emph{Time Series Classification} (TSC) \cite{bagnall2017great} and \emph{Time Series Forecasting} (TSF) \cite{hyndman2018brief,hyndman2008forecasting}.
Research in TSC has greatly benefited from the University of California Riverside and University of East Anglia (UCR/UEA) Time Series Archives \cite{dau2019ucr,bagnall2018uea}.
The univariate TSC archive was first released in 2002 with 16 datasets to encourage a more rigorous evaluation of TSC algorithms \cite{keogh2003need}.
In 2015, it was expanded to 85 datasets, covering a wider range of problems.
Then, it was criticised of not being a good representative of the real-world problems where time series often have missing values and are of varying lengths.
Hence, the archive was recently expanded to 128 datasets that now include time series of varying lengths, un-normalised time series and time series with missing values \cite{dau2019ucr}.
The first official multivariate TSC archive \cite{bagnall2018uea} was recently released by researchers from UEA.
It contains 30 multivariate time series datasets of equal length with no missing values.  
Previously, there were only 12 small multivariate time series datasets from Baydogan \cite{baydogan2015learning}. 
These archives have motivated the development of numerous new state-of-the-art TSC algorithms in the last five years \cite{lines2015time,bagnall2015time,lines2016hive,lucas2019proximity,shifaz2020ts,fawaz2019deep,dempster2020rocket}, 
each of them being more accurate than their predecessors. 

On the other hand, the advancement in \emph{Time Series Forecasting} relies on time series forecasting competitions \cite{hyndman2018brief}. 
The most popular ones being the Makridakis competitions, also known as the M-competitions.
The M-competitions were started by Spyros Makridakis and Michèle Hibon \cite{hyndman2018brief,makridakis1982accuracy,makridakis1993m2,makridakis2000m3,makridakis2018m4,makridakis2020m4}.
They were the first few researchers who put together 111 time series to compare different forecasting methods \cite{hyndman2018brief}. 
In 1982, they held the first M-competition involving 1001 time series, comparing 15 TSF algorithms \cite{makridakis1982accuracy}.
This competition motivated researchers to focus more on algorithms that give good forecasts and treat TSF a different problem from time series analysis \cite{hyndman2018brief,makridakis1982accuracy}.
The competition continues until today with 3003 time series for the M3 competition \cite{makridakis2000m3} and 100,000 time series in the recent M4 competition \cite{makridakis2018m4,makridakis2020m4}.
Both of these competitions involved time series of varying lengths, taken from business, demography, finance and economics.
Besides, there are also other competitions such as NN3 and NN5 Neural Network competitions \cite{crone2011advances} and a few Kaggle competitions \cite{athanasopoulos2011tourism,web_traffic_forecast}.

Each year, thousands of papers proposing new algorithms for (a) TSC -- to predict a discrete label of the time series and (b) TSF -- to predict some continuous values of a time series in the future using recent and seasonal values, have utilised the benchmarking archives mentioned earlier. 
These algorithms are designed for these specific problems, but may not be useful for other problems.
For example, TSC and TSF methods are not suitable if we wish to predict the heart rate of a person using photoplethysmogram (PPG) and accelerometer data \cite{reiss2019deep}, which is a continuous value, not a future value and does not depend on the recent PPG data. 

We refer to this problem as \emph{Time Series Extrinsic Regression} (TSER).
Not to be confused with the TSF community where the term \emph{Time Series Regression} usually means fitting the historical time series data with a regression model, such as Autoregression (\arima{}) \cite{box1970time} or Exponential Smoothing \cite{gardner1985exponential,hyndman2008forecasting}.
These algorithms fit to recent and/or seasonal values of the time series and extrapolate to forecast future values.
Here, we are interested in a more general methodology of predicting a single continuous value, from univariate or multivariate time series.
We aim to predict values that can be either a continuation of the input time series or external to it and do not necessarily need to be a future value or depend on recent heavily on recent values.
In the case where predicting a future value is of interest, then that becomes a TSF problem and if predicting a discrete value is of interest, then that becomes a TSC problem.

To the best of our knowledge, 
research into TSER has received much less attention in the time series research community and there are no algorithms developed for general time series extrinsic regression problems.
Most algorithms are developed for a specific problem \cite{reiss2019deep,zhang2015photoplethysmography,zhang2014troika}.
In the machine learning (ML) research community, this is commonly referred to as ``regression'', where a single continuous value is predicted from a set of features \cite{sammut2011encyclopedia}. 
These features are derived from the data and are usually not correlated to each other or related in time.
Features that are highly correlated are typically being treated as redundant, i.e. only one of them is sufficient to achieve similar performance. 
In our context, these features are time series (a sequence of values) instead of a single value.
TSER can also be considered a special case of \emph{scalar-on-function regression (SoFR)} from the statistics community \cite{reiss2017methods, goldsmith2014estimator}. 
SoFR considers a time series as functional data and builds statistical models to map functional data to a scalar response value. 
In our case, we address the problem from a ML perspective, treating it as a regression problem.

We aim to motivate and support the research into TSER by introducing the first TSER benchmarking archive.
This archive contains 19 datasets from different domains, with varying number of dimensions, unequal length dimensions and missing values. 
The rest of this paper is organised as follows.
In Section \ref{sec:datasets}, we describe the datasets that are in the archive.
Section \ref{sec:baseline} sets a baseline to the datasets by adapting state-of-the-art TSC and machine learning regression algorithms. 
Finally, in Section \ref{sec:conclusion}, we summarise our contribution and give some direction for future work.

\section{Datasets}
\label{sec:datasets}
This section outlines the datasets in this TSER archive. 
The current archive contains 19 time series datasets as shown in Table \ref{tab:datasets}.
They are available online at \url{http://timeseriesregression.org/}.
The archive contains 8 datasets adapted from the UCI machine learning repository \cite{Dua:2019}, 3 from Physionet, 1 from a signal processing competition \cite{zhang2014troika}, 1 from the World Health Organisation (WHO), 1 from the Australian Bureau of Meteorology (BOM) and the rest are donations.  

\begin{table}[]
    \centering
    \resizebox{\textwidth}{!}{
    \begin{tabular}{|l|l|l|c|c|c|c|c|c|}
        \hline
        & Type & Dataset & Train size & Test size & Length & No of Dimension & Missing \\
        \hline
        1 & Energy Monitoring & AppliancesEnergy & 96 & 42 & 144 & 24 & No \\ 
        2 & Energy Monitoring & HouseholdPowerConsumption1 & 746 & 694 & 1440 & 5 & Yes \\
        3 & Energy Monitoring & HouseholdPowerConsumption2 & 746 & 694 & 1440 & 5 & Yes \\
        \hline 
        4 & Environment Monitoring & BenzeneConcentration & 3433 & 5445 & 240 & 9 & Yes \\
        5 & Environment Monitoring & BeijingPM25Quality & 12432 & 5100 & 24 & 9 & Yes \\
        6 & Environment Monitoring & BeijingPM10Quality & 12432 & 5100 & 24 & 9 & Yes \\
        7 & Environment Monitoring & LiveFuelMoistureContent & 3493 & 1510 & 365 & 7 & No \\
        8 & Environment Monitoring & FloodModeling1 & 471 & 202 & 266 & 1 & No \\
        9 & Environment Monitoring & FloodModeling2 & 389 & 167 & 266 & 1 & No \\
        10 & Environment Monitoring & FloodModeling3 & 429 & 184 & 266 & 1 & No \\
        11 & Environment Monitoring & AustraliaRainfall & 112186 & 48081 & 24 & 3 & No \\
        \hline
        12 & Health Monitoring & PPGDalia* & 43215 & 21482 & 256,512 & 4 & No \\ 
        13 & Health Monitoring & IEEEPPG & 1768 & 1328 & 1000 & 5 & No \\
        14 & Health Monitoring & BIDMCRR & 5471 & 2399 & 4000 & 2 & No \\
        15 & Health Monitoring & BIDMCHR & 5550 & 2399 & 4000 & 2 & No \\
        16 & Health Monitoring & BIDMCSpO2 & 5550 & 2399 & 4000 & 2 & No \\
        \hline 
        17 & Sentiment Analysis & NewsHeadlineSentiment & 58213 & 24951 & 144 & 3 & No \\
        18 & Sentiment Analysis & NewsTitleSentiment & 58213 & 24951 & 144 & 3 & No \\
        \hline
        19 & Forecasting & Covid3Month & 140 & 61 & 84 & 1 & No \\
        \hline
    \end{tabular}
    }
    \caption{Time series datasets in the current TSER archive. The ones marked with an asterisk (*) have different lengths between the dimensions but are still equal in length between all the instances in the dataset.}
    \label{tab:datasets}
\end{table}

This archive currently covers 5 application areas, \emph{Energy Monitoring}, \emph{Environment Monitoring}, \emph{Health Monitoring}, \emph{Sentiment Analysis} and  \emph{Forecasting}.
The datasets are formatted with the \texttt{.ts}\footnote{\url{https://alan-turing-institute.github.io/sktime/examples/loading_data.html}\label{fnote: loading ts}} format used in tsml\footnote{\url{https://github.com/uea-machine-learning/tsml}} and sktime\footnote{\url{https://github.com/alan-turing-institute/sktime}} time series machine learning repositories.  
An example of loading the data into Python can be found on the sktime website\footref{fnote: loading ts} and our github page\footnote{\url{https://github.com/ChangWeiTan/TSRegression}}.
Missing values in the original dataset were not imputed and represented by the `?' symbol, following the \texttt{.ts} convention used in the UCR/UEA archives \cite{dau2019ucr,bagnall2018uea}. 
For fair comparison of regression algorithms, we split the datasets in the archive into predefined train and test sets which will be outlined in the following sections. 

\subsection{Energy monitoring}
Energy monitoring monitors the energy usage of a building by collecting various data such as temperature, humidity, rain, voltage and current readings from sensors attached all over a building.
These data are collected in the form of time series and are mapped to the power consumption of the building.
For example, higher power consumption will be observed when the temperature is low, during winter months as more energy is required to heat up a building. 
They are then used to optimise the energy usage which can save millions of dollars for a large building. 
In this section, we explain three datasets for \emph{energy monitoring} obtained from two sources.

\subsubsection{Appliances energy prediction}
\begin{figure}[!h]
	\centering
	\begin{subfigure}{0.49\linewidth}
		\includegraphics[width=\linewidth]{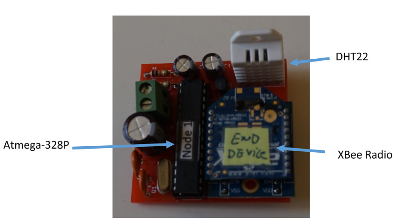}
		\caption{}
		\label{fig:appliances energy sensor}
	\end{subfigure}
	\begin{subfigure}{0.49\linewidth}
		\includegraphics[width=\linewidth]{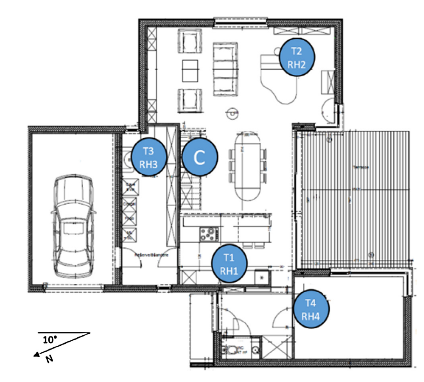}		\caption{}
		\label{fig:appliances energy house}
	\end{subfigure}
	\caption{(a) The Zigbee wireless sensor and (b) house layout for collecting the temperature and humidity of the house. Figures taken from \cite{candanedo2017data} with permission from the authors.}
\end{figure}

Luis \etal{} \cite{candanedo2017data} studied algorithms for predicting the energy usage of appliances.
The authors monitored the temperature and humidity of different rooms in a house for 4.5 months using ZigBee wireless sensor network, illustrated in Figure \ref{fig:appliances energy sensor}.
They measured the temperature and humidity of the kitchen, living, laundry, office, bathroom, ironing, 2 bedrooms and outside of the house.
Figure \ref{fig:appliances energy house} shows an example of this layout.
Weather data from a nearby airport station, Chievres Airport, Belgium were also being used to improve the predictions.
The ground truth, appliances energy, was recorded with m-bus energy meters at 10 minutes interval. 
Data filtering and feature ranking techniques were discussed in the paper to remove non-predictive parameters \cite{candanedo2017data}.
Then, four algorithms, (1) multiple linear regression, (2) support vector machine with radial kernel, (3) random forest and (4) gradient boosting machine (GBM) were evaluated on the collected data. 
Although the results showed that GBM performed the best, it was only able to explain 57\% of the variance (R2) in the test set. 
This implies the need for better predictive algorithms.

\begin{figure}[!h]
    \centering
    \includegraphics[width=0.9\columnwidth]{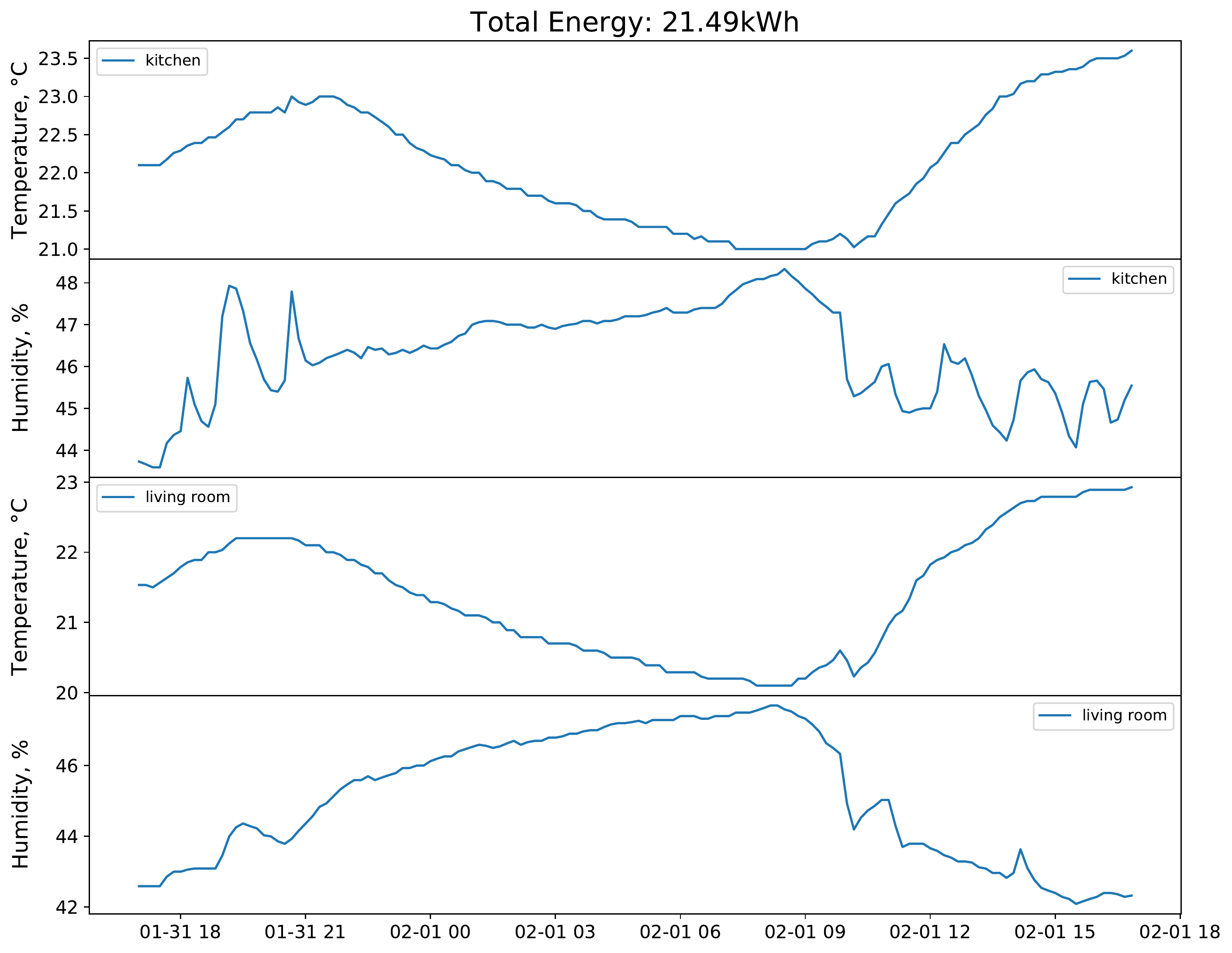}
    \caption{Example of measurements taken at 10 minutes interval from wireless sensors used to monitor the temperature and humidity of a house.}
    \label{fig:appliances energy}
\end{figure}

We created the \textbf{AppliancesEnergy} dataset using this dataset and reformulated the problem. 
The goal of the original paper \cite{candanedo2017data} was to predict the instantaneous energy usage given all the sensor measurements at a given time point. 
Instead, we reformulated the problem as given the daily time series of each sensor's measurement measured at every 10 minutes interval, predict the total daily appliances energy in kWh. 
This forms a time series with 144 data points per day as shown in Figure \ref{fig:appliances energy}.
Each time series in our AppliancesEnergy dataset consists of 24 variables.
The variables are the temperature and humidity measurements of each room in the house and the weather data obtained from the nearby airport.
This dataset is split into train and test sets by randomly sampling 70\% as train and the remaining 30\% as test. 
This results in 96 train time series and 42 test time series.

\subsubsection{Individual household electric power consumption}
\begin{figure}[!h]
    \centering
    \includegraphics[width=0.8\columnwidth]{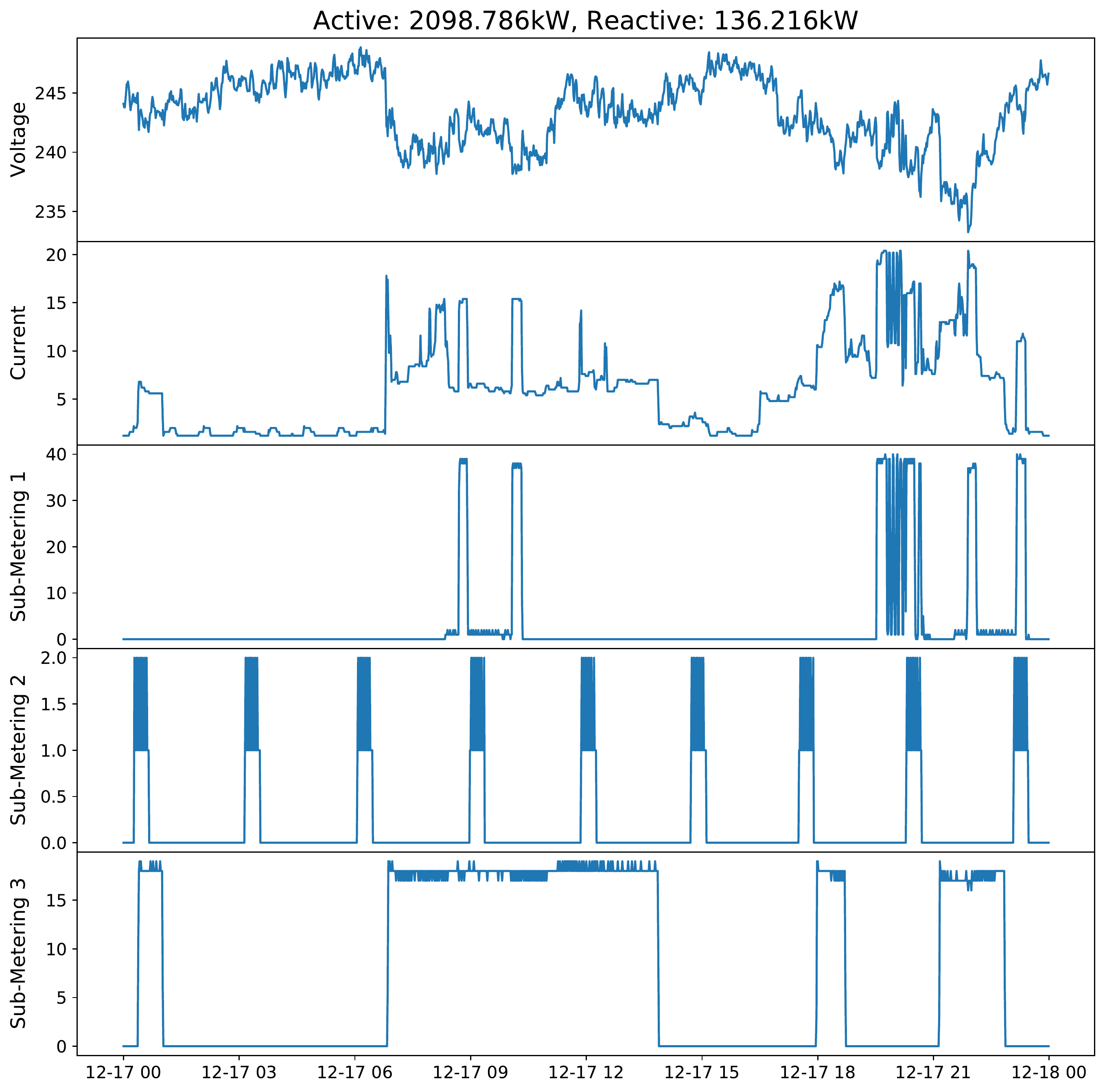}
    \caption{Examples of the daily voltage, current and sub-metering measurements used to predict the total daily active and reactive power consumption in a house.}
    \label{fig:household power}
\end{figure}

This dataset was sourced from the UCI repository\footnote{\url{https://archive.ics.uci.edu/ml/datasets/Individual+household+electric+power+consumption}} \cite{Dua:2019}. 
It contains 2 million measurements gathered over a period of 47 months, between December 2006 and November 2010. 
The data measures the minutely global active and reactive power, voltage, current and sub-meter energy of a house located in Sceaux, 7km of Paris.
This dataset is converted into a TSER problem for this archive, using the daily voltage, current and 3 sub-metering data to predict the total daily active and reactive power consumptions.
Figure \ref{fig:household power} shows an example of such time series used for the prediction with the five variables and a length of 1440.
The \textbf{HouseholdPowerConsumption1} and \textbf{HouseholdPowerConsumption2} datasets represent the dataset for active and reactive power prediction, respectively. 
The datasets are split into train and test sets by taking all measurements before 2009 as train and after 2009 as test. 
Both datasets have 746 and 694 train and test time series, respectively.

\subsection{Environment monitoring}
Environment monitoring has become more important than ever with climate change getting more serious.
It is the task to predict anything related to our environment such as pollution level, rainfall, crop yield and flood water level.
This section outlines eight \emph{environment monitoring} datasets in this archive, obtained from four sources.

\subsubsection{Air quality}
One of the main applications of environment monitoring is to predict air quality for pollution monitoring. 
Vito \etal{} studied the calibration of chemical sensors for benzene estimation to monitor air pollution in an Italian city \cite{de2008field}.
They used five metal oxide chemical sensors embedded in an air quality chemical multisensor device to record the hourly air pollutant concentrations from March 2004 to February 2005 \cite{de2008field}.
The ground truth concentrations for 5 atmospheric pollutants, Carbon Monoxide, Non Metanic Hydrocarbons, Benzene, Nitrogen Oxides and Nitrogen Dioxide was obtained from a fixed weather station.
Besides the pollutants, local temperature, relative humidity and absolute humidity data were also recorded. 

\begin{figure}[!h]
    \centering
    \includegraphics[width=0.9\columnwidth]{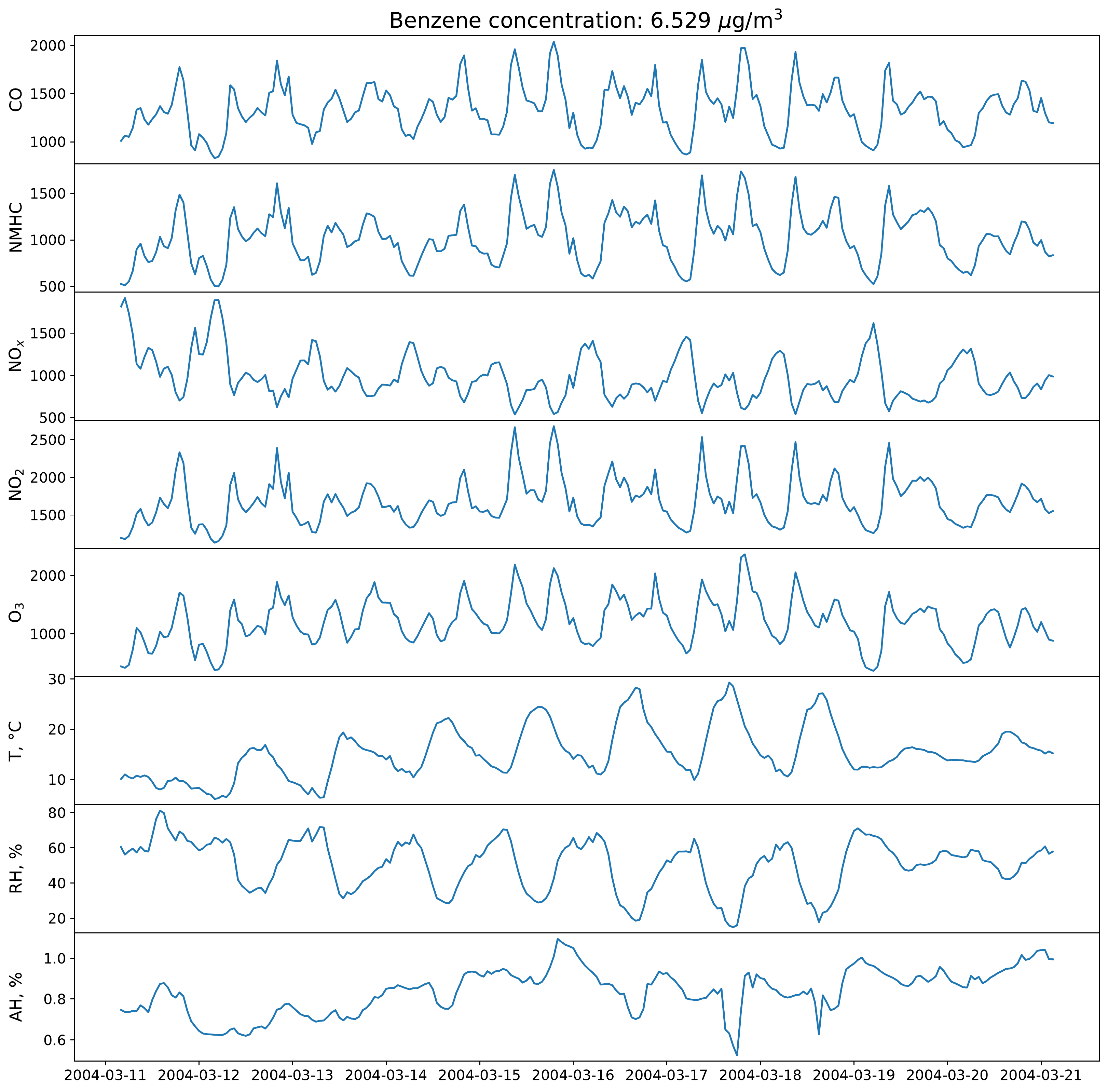}
    \caption{Example of measurements from the chemical sensors and weather data used to estimate benzene concentration in an Italian city.}
    \label{fig:benzene concentration}
\end{figure}

We created the \textbf{BenzeneConcentration} dataset using the dataset provided in the study \cite{de2008field}.
Apart from the five chemical sensors, this dataset also uses the temperature, relative humidity and absolute humidity data, forming an 8-dimensional time series dataset.  
As the data was originally used for calibrating chemical sensors, we formulate the regression problem as predicting the benzene concentration for the current hour using the hourly measurements from the last 10 days, forming a time series of length 240.
The 10-days window was found to give good calibration results from the paper \cite{de2008field}. 
Note that the 10-day segment will not be used if the benzene concentration for the current hour is missing. 
Figure \ref{fig:benzene concentration} shows an example of the time series measurements used to estimate benzene concentration in the Italian city.
The training set consists of the initial 8 months data while the remaining are used as test set.
This results in 3433 training instances and 5445 test instances.  

\subsubsection{Beijing multi-site air quality}
The capital of China, Beijing, is one of the cities in the world with the worst air pollution levels. 
Numerous studies have been conducted to study and reduce the air pollution level in Beijing \cite{zhang2017cautionary}.
In 2017, the Beijing Municipal Environmental Monitoring Center (BMEMC) reported a reduction of 9.9\% in fine particulate matter (PM2.5) level from the previous year \cite{zhang2017cautionary}.
However, a study conducted by Zhang \etal{} \cite{zhang2017cautionary} shows that there was uncertainty in the report, as they studied the past 4 years of Beijing's PM2.5 and PM10 data at 36 monitoring sites. 
Hourly air pollutants data such as SO$_2$, NO$_2$, CO and O$_3$ concentrations as well as meteorological data from the air quality monitoring sites were used in the study.  

\begin{figure}[!h]
    \centering
    \includegraphics[width=0.9\columnwidth]{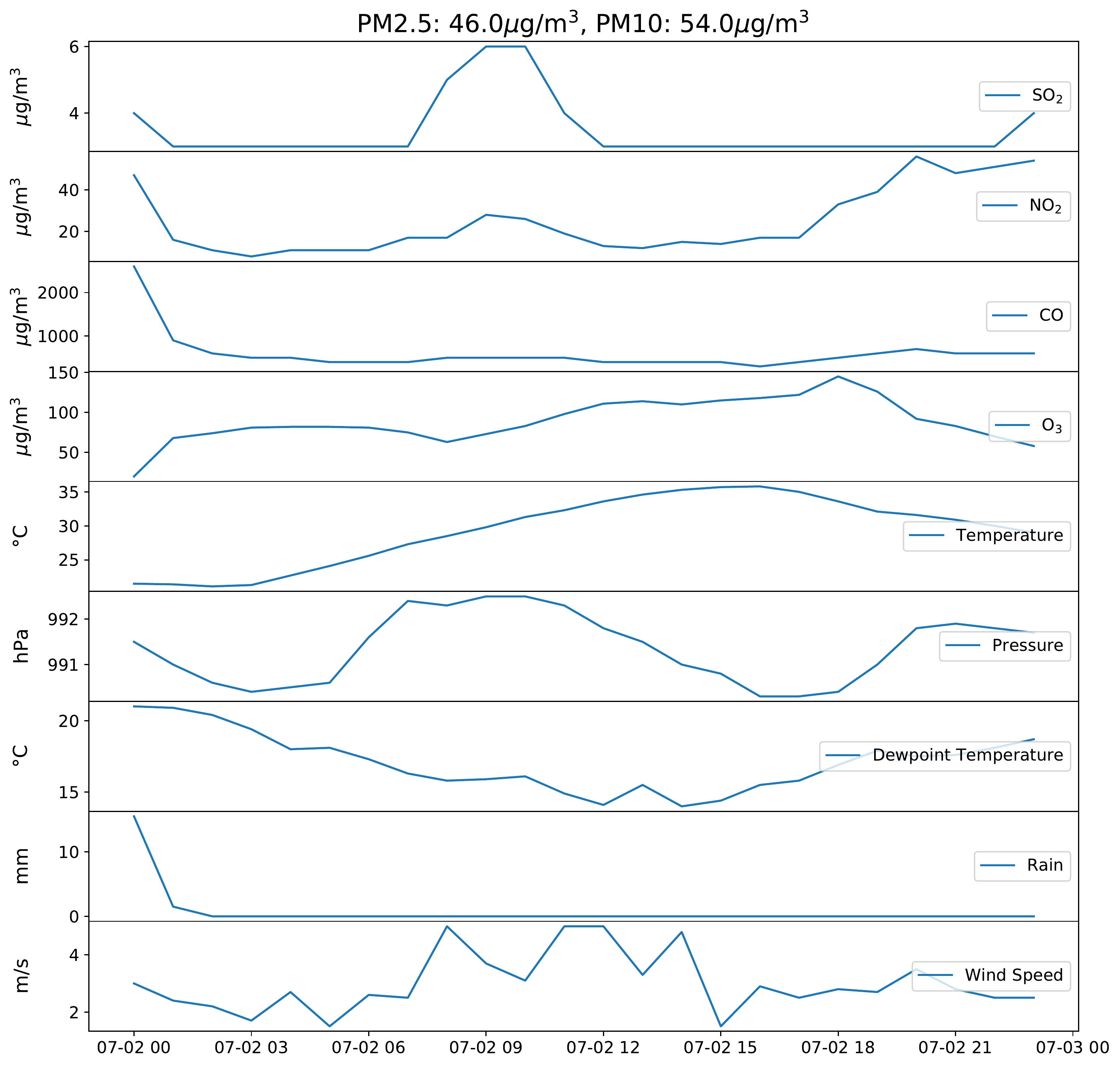}
    \caption{Example of a 9-dimensional time series measuring the four daily air pollutants as well as five weather data measurements to predict the PM2.5 and PM10 level in the city of Beijing.}
    \label{fig:beijing}
\end{figure}

The \textbf{BeijingPm25Quality} and \textbf{BeijingPm10Quality} datasets are created from the dataset provided by \cite{zhang2017cautionary}.
The goal is to predict both PM2.5 and PM10 level of Beijing using 9-dimensional time series that measures the four daily air pollutants as well as five meteorological data (temperature, dewpoint temperature, wind speed, pressure and rain amount) from 12 air-quality monitoring sites in Beijing.
Figure \ref{fig:beijing} shows an example of the 9-dimensional time series in these datasets used to predict the PM2.5 and PM10 level in the city of Beijing.
The training set consists of all time series taken before the year 2016, with a total of 12432 time series.
The test set contains 5100 time series which consists of measurements taken after the year 2016.

\subsubsection{Live fuel moisture content}
Apart from pollution monitoring, bush fire monitoring is also an important application of \emph{environment monitoring}.   
One way to monitor bush fire is to monitor the moisture in the vegetation, i.e. the ratio between the weight of water in vegetation and the weight of the dry part of vegetation (information that is obtained by sampling vegetation in the field, weighing it and drying it to weigh it again). 
This is known as the live fuel moisture content (LFMC) and is an important variable as the risk of fire increases very rapidly as soon as the LFMC goes below 80\% \cite{yebra2018fuel}. 
We have obtained a LFMC database from researchers at Monash University who are working on developing models to predict LFMC values.
They used the Globe-LFMC dataset as the ground truth. 
Globe-LFMC is an extensive global database of LFMC containing 161,717 instances, measured from 1383 sampling sites in 11 countries. 
One year of daily reflectance data at 7 spectral bands (459 nm to 2155 nm) before the LFMC sampling date from the Moderate Resolution Imaging Spectrometer (MODIS satellite are one of the inputs to their model.
The elevation, slope and aspect of the sampling site extracted from the Advanced Spaceborne Thermal Emission and Reflection Radiometer (ASTER) Global Digital Elevation Model Version 3 (GDEM 003) are also being considered.
\vspace{10pt}


\begin{figure}[!h]
	\centering
	\includegraphics[width=\linewidth]{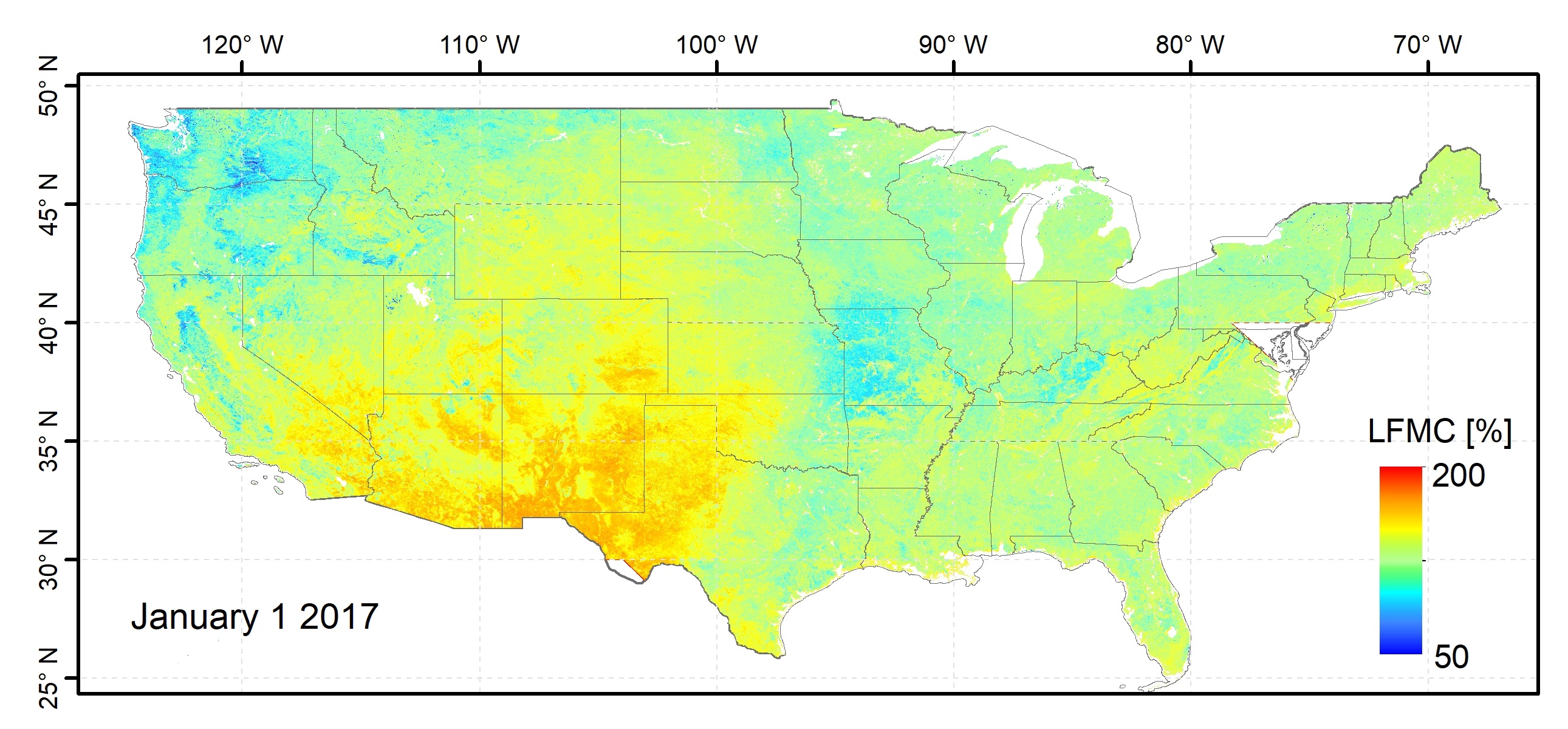}		
	\caption{An example of LFMC map for the United States.}	
	\label{fig:retrieved lfmc}
\end{figure}

\begin{figure}[!h]
    \centering
    \includegraphics[width=0.8\columnwidth]{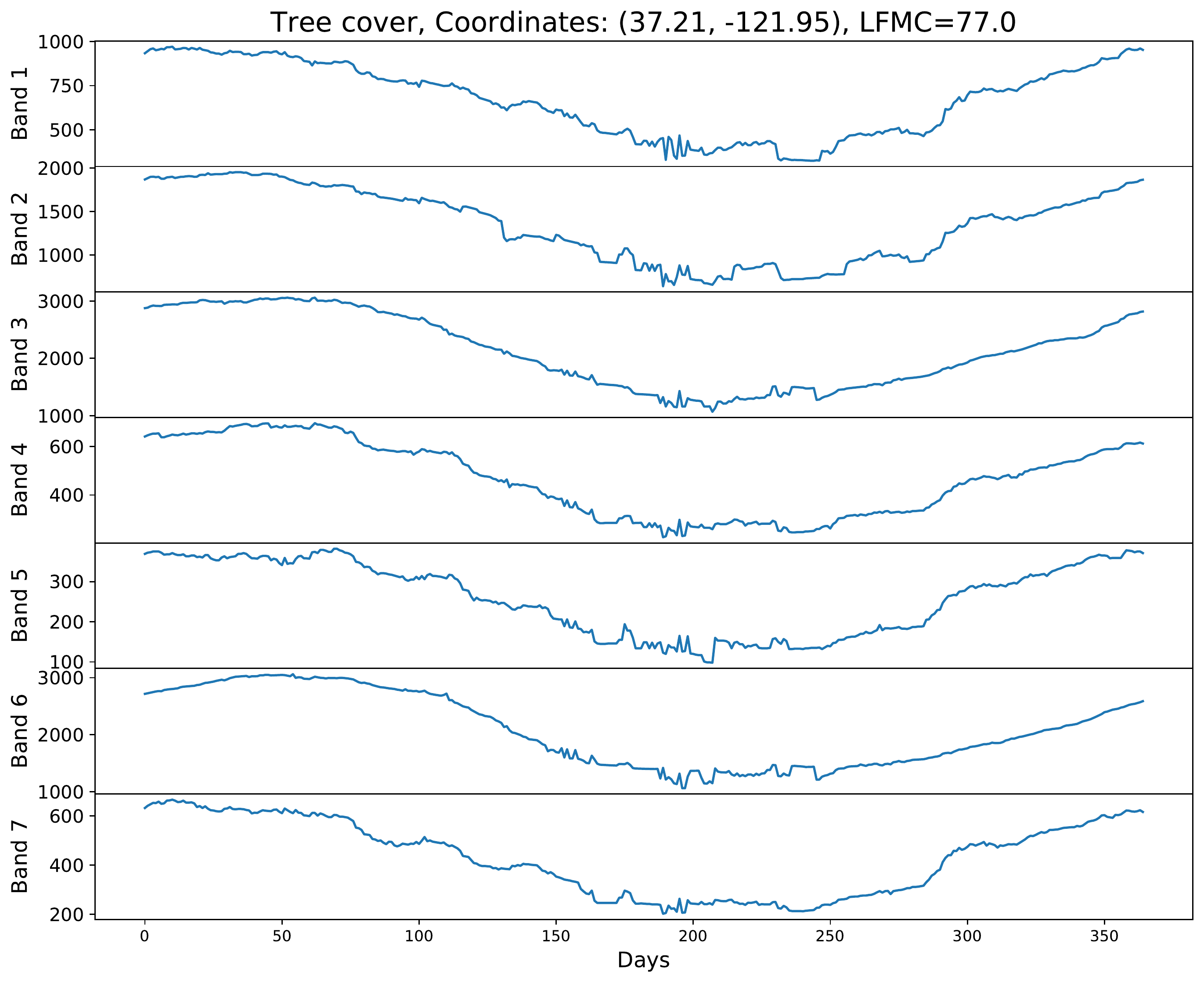}
    \caption{Example of LFMC time series with 7 spectral bands.}
    \label{fig:lfmc}
\end{figure}

We created the \textbf{LiveFuelMoistureContent} dataset from the database by stratified sampling 5003 instances across the United States.
Figure \ref{fig:retrieved lfmc} shows an example of the retrieved LFMC values for the United states. 
Stratify sampling ensures that the land cover classes in the dataset are well distributed and balanced.
Then 70\% of the dataset is randomly selected as the training set with 3493 time series and 1510 time series in the test set.
Figure \ref{fig:lfmc} illustrates the time series with 7 spectral bands in the dataset.

\subsubsection{Flood modeling}
\begin{figure}[!h]
	\centering
	\begin{subfigure}{0.49\linewidth}
		\includegraphics[width=\linewidth]{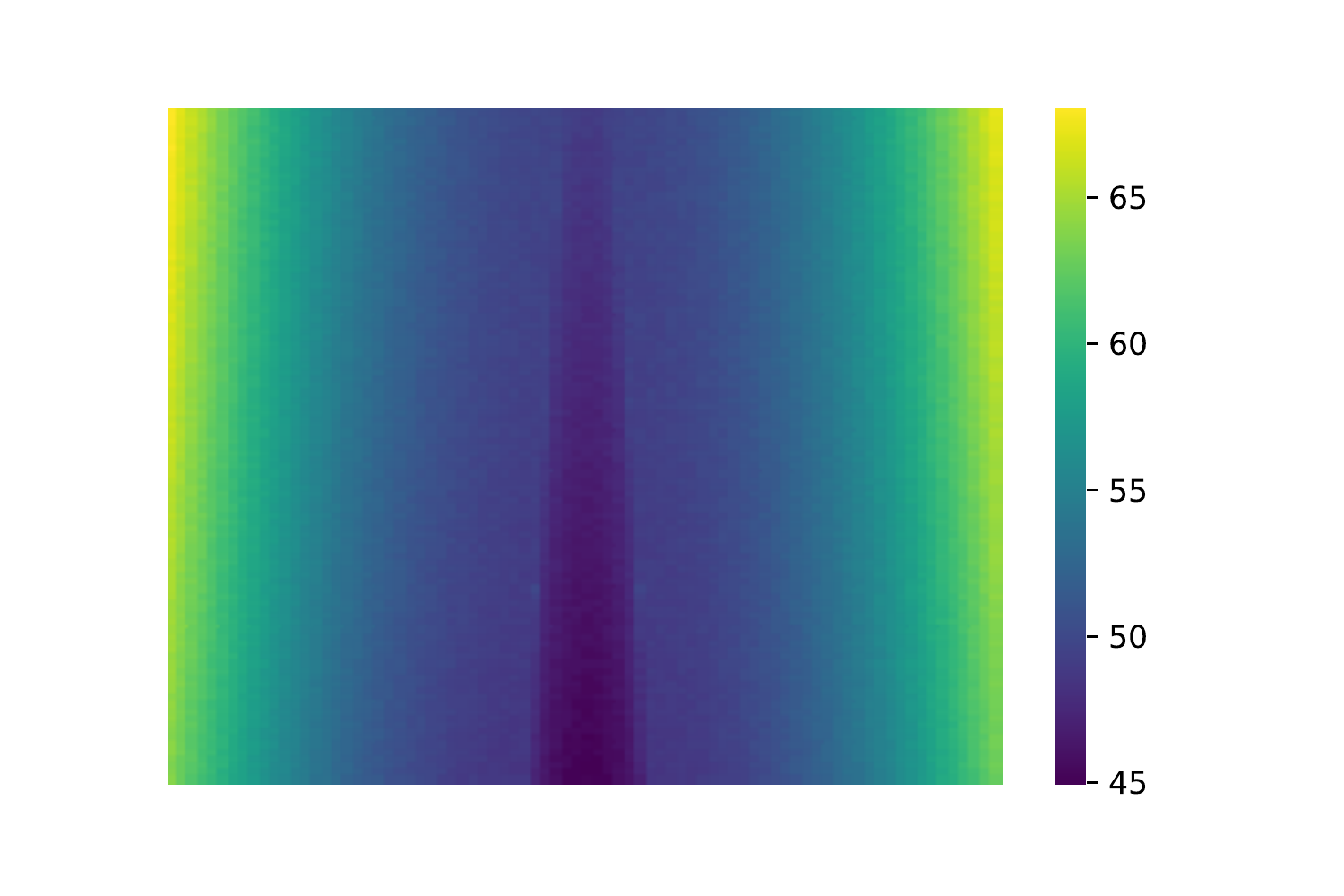}
		\caption{}
		\label{fig:dem}
	\end{subfigure}
	\begin{subfigure}{0.49\linewidth}
		\includegraphics[width=\linewidth]{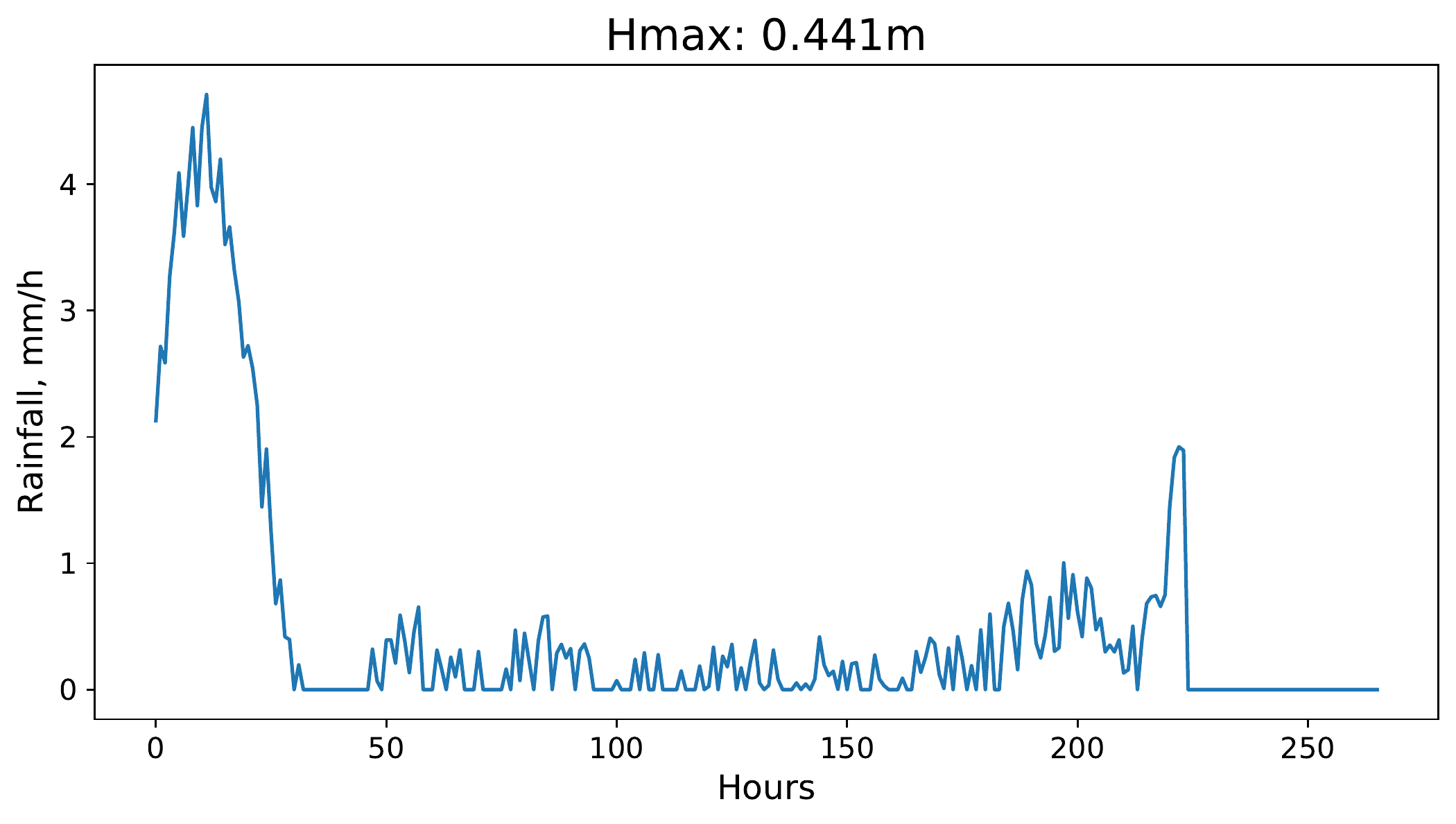}		\caption{}
		\label{fig:rainfall flood}
	\end{subfigure}
	\caption{Example of (a) a synthetic Digital Elevation Model (DEM) topography and (b) a rainfall event that leads to maximum water depth, Hmax that is near the outlet of the DEM.}
\end{figure}

Flood modelling consists of solving simplified fluid dynamics equations over a given domain or DEM (Digital Elevation Model) in response to a rainfall time series event.
Different processes can happen once the precipitation reaches the ground: infiltration, evaporation, transpiration, interaction with a pipe network or hydraulic structures, and what is left is called surface runoff. 
Runoff describes the path of water following the slope of the terrain and ending up in a stream.
Typically, the input rainfall time series and the 2D DEM topography is passed into a software (Lisflood-FP) that solves the fluid dynamics equations and outputs the water flow (m$^3$/s) and water depth (in m). 
In flood studies, researchers are mostly interested in knowing if, where and how much the domain will get flooded, and produce flood maps.
A flood map is a distributed view of the maximum water depth reached due to a given rainfall event.
Figure \ref{fig:dem} and \ref{fig:rainfall flood} show an example of a simple synthetic DEM with a water stream in the middle and a rainfall event that leads to a maximum water depth of 0.441m, respectively.

For the archive, we obtained three synthetic DEMs and rainfall events from flood studies researchers at Monash University. 
Synthetic data were used because real DEM data that pairs with accurate rainfall events are rare.
These DEMs consist of a square grid with different types of terrains and a water stream in the middle of the DEM as shown in Figure \ref{fig:dem}.
Then we have the rainfall time series event for the DEM, illustrated in Figure \ref{fig:rainfall flood} which gives the maximum water height near the outlet of the DEM.
We created the \textbf{FloodModeling1}, \textbf{FloodModeling2} and \textbf{FloodModeling3} datasets from these synthetic DEMS, each of them having a different number of rainfall events.
All three datasets are split into train and test set by randomly sampling 70\% as the training set. 
FloodModeling1 has 471 training and 202 test time series.
FloodModeling2 has 389 training and 167 test time series.
FloodModeling3 has 429 training and 184 test time series.

\subsubsection{Rainfall predictions}
\begin{figure}[!h]
    \centering
    \includegraphics[width=0.9\columnwidth]{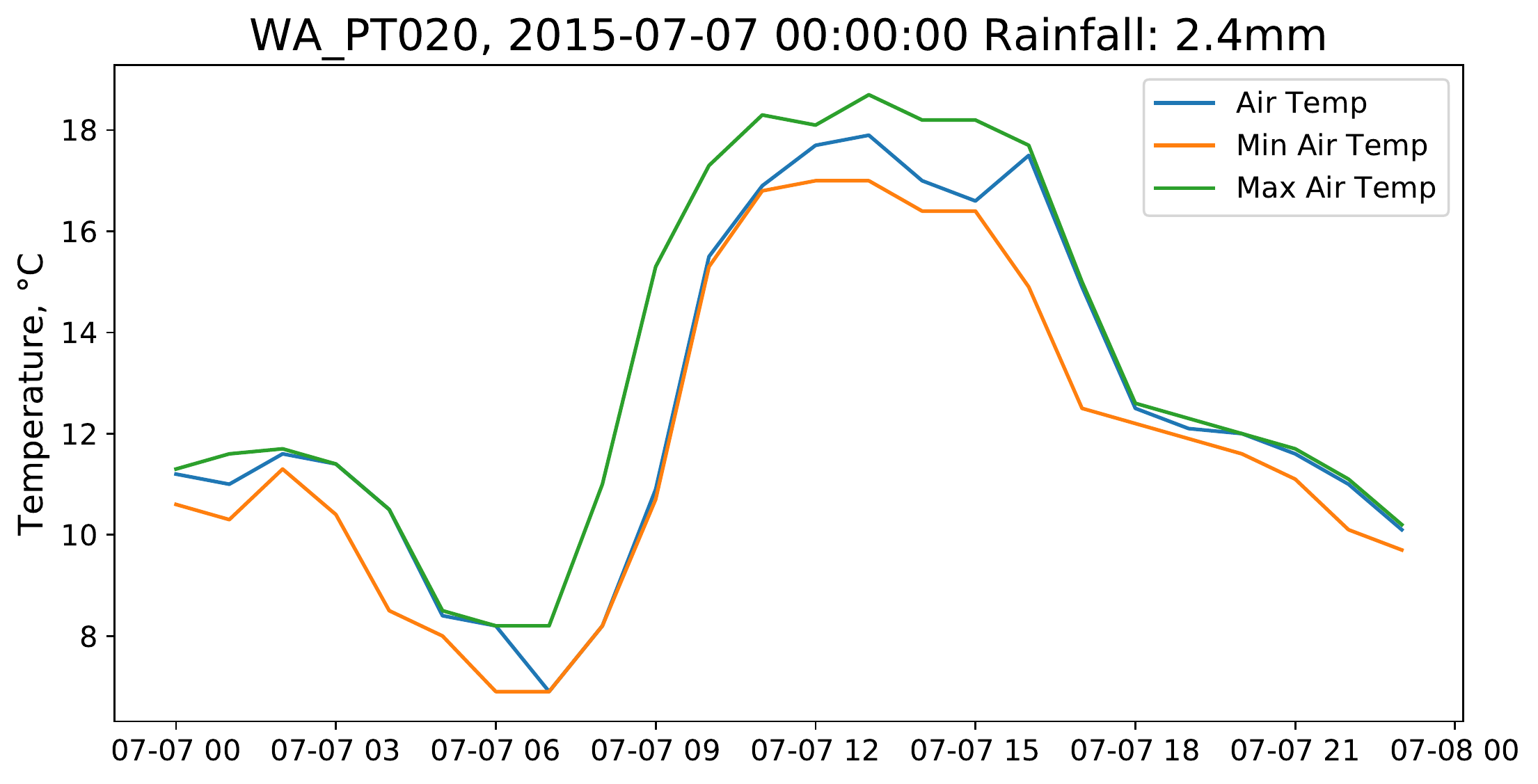}
    \caption{Example of air temperature measurements taken from one of the Western Australian weather stations for rainfall prediction.}
    \label{fig:australia rainfall}
\end{figure}

An important task in \emph{environment monitoring} is to predict rainfall.
The Australian Bureau of Meteorology (BOM) released a dataset\footnote{\url{https://data.gov.au/data/dataset/weather-forecasting-verification-data-2015-05-to-2016-04}} that contains a year of temperature and rainfall data from May 2015 to April 2016.
The dataset was collected from 518 weather stations throughout all of Australia. 
This dataset was aggregated into hourly values and was used for the comparison and verification of temperature and rainfall forecasts.
The dataset contains the hourly average, maximum and minimum temperature as well as the rainfall amounts.
We adapted this dataset to create the \textbf{AustraliaRainfall} dataset to predict the total daily rainfall using 24 hours of temperature measurements.
Figure \ref{fig:australia rainfall} shows an example of the air temperature measured from a Western Australia weather station and the total daily rainfall. 
The dataset is split into training and test sets by randomly sampling 70\% as the training set.
There are 112186 and 48081 time series in train and test sets respectively.

\subsection{Health monitoring}
Health monitoring is the task of monitoring the health or vital signs of an individual using some devices.
For example, estimating the heart rate, respiratory rate and blood oxygen saturation level.
The data typically comes from a wearable device that can be attached to the subject, such as a photoplethysmogram (PPG), electrocardiogram (ECG), electroencephalogram (EEG) or accelerometers but could also come from medical devices.
In this section, we describe five \emph{health monitoring} datasets that come from three sources.
These datasets focus on three tasks, estimating heart rate, respiratory rate and blood oxygen saturation level. 

\subsubsection{PPG-DaLiA}
PPG sensors are now widely used in many smart wearable devices such as the Fitbit and Apple Watch to measure heart rate \cite{reiss2019deep}.
Although ECG is more precise in determining the heart rate, it is cumbersome in daily life settings \cite{reiss2019deep}. 
PPG-based heart rate estimation is still a challenging task \cite{reiss2019deep}. 
Previous methods of estimating the heart rate from PPG sensors mostly relies on spectral analysis \cite{zhang2014troika,zhang2015photoplethysmography,salehizadeh2016novel,schack2017computationally} and they are not very accurate \cite{reiss2019deep}.
The authors from \cite{reiss2019deep} proposed a convolutional neural network based approach that takes the signal in the frequency domain as input.
They showed that their approach is significantly more accurate compared to the existing spectral methods.

\begin{figure}[!h]
    \centering
    \includegraphics[width=0.65\columnwidth]{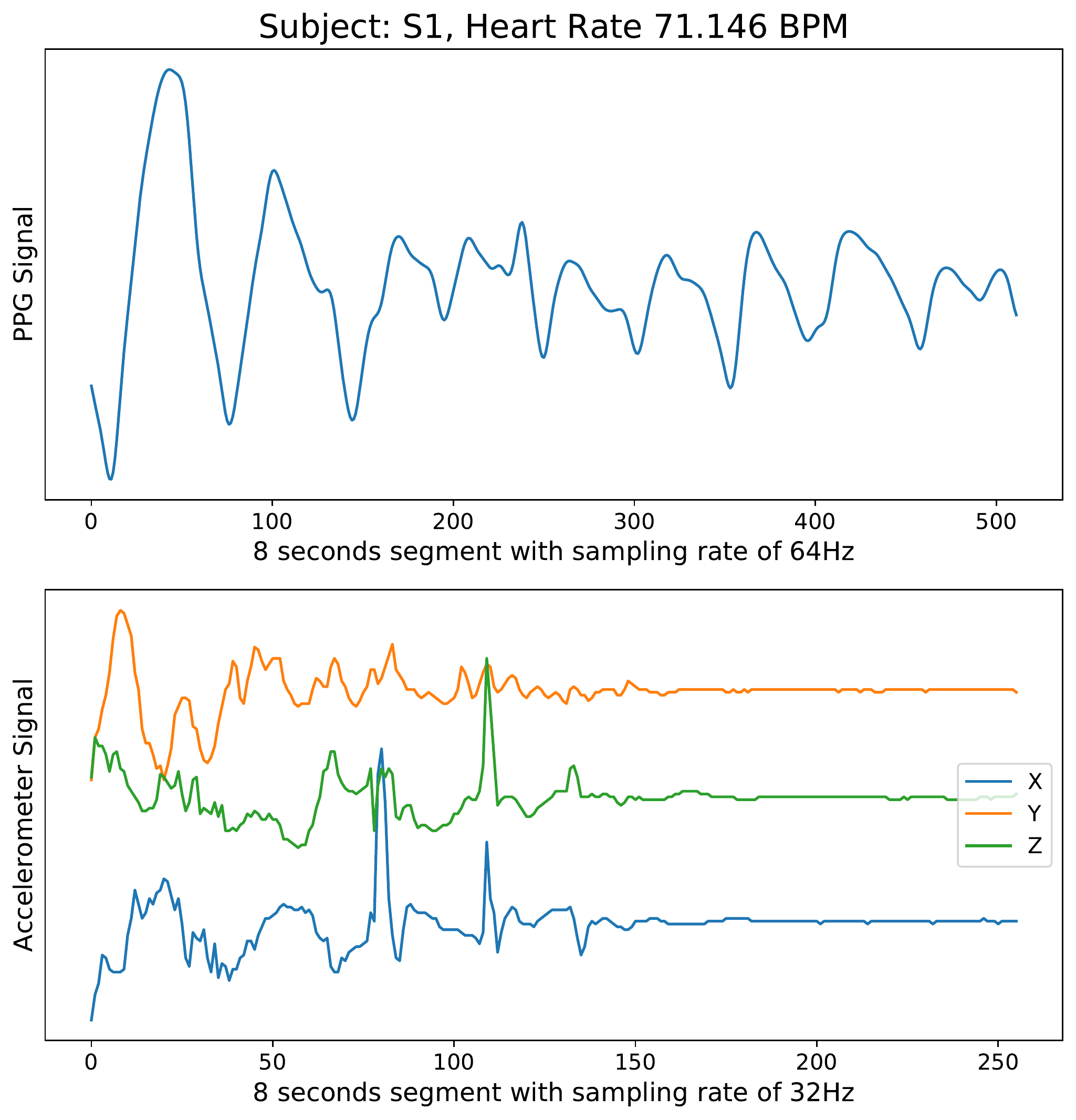}
    \caption{Example of time series in the PPGDalia dataset. The title shows the subject and the current heart rate in beats per minute (BPM). Note the different sampling rate in both signals.}
    \label{fig:ppg_dalia}
\end{figure}

We adapted the original PPG-DaLiA dataset from the UCI machine learning repository \cite{Dua:2019} and created the \textbf{PPGDalia} dataset for our TSER archive. 
PPG-DaLiA contains a single channel PPG and 3D accelerometer motion data recorded from 15 subjects performing a wide range of real-life activities, creating a 4 dimensional time series.
The measurements from each subject are then segmented into 8 seconds windows with 6 seconds overlaps \cite{reiss2019deep}.
In \textbf{PPGDalia}, subjects 1 to 10 are selected to be in the training set and the remaining are in the test set, resulting in 43215 train instances and 21482 test instances.
Figure \ref{fig:ppg_dalia} shows an example of the PPG signal and accelerometer signals in the dataset.
Note that the time series in the PPG and accelerometer channels have different lengths due to different sampling rate of 64Hz and 32Hz respectively.

\subsubsection{IEEE Signal Processing Cup 2015}
In 2015, IEEE organised a signal processing competition\footnote{IEEE Signal Processing Cup 2015: Heart Rate Monitoring During Physical Exercise Using Wrist-Type Photoplethysmographic (PPG) Signals} to monitor heart rate using wrist type PPG signals \cite{zhang2014troika}, similar to Apple Watch in Figure \ref{fig:apple watch}.
They released a dataset that contains 2 channel PPG signals, 3-axis acceleration signals and 1 channel ECG signals, all sampled at 125Hz.
The dataset was recorded from 12 subjects aged between 18-35 years old, running on a treadmill with changing speeds 
\cite{zhang2014troika}.
The owner of the dataset proposed a spectral analysis method to estimate heart rate from PPG signals \cite{zhang2014troika}.

\begin{figure}[!h]
    \centering
    \includegraphics[width=0.8\columnwidth]{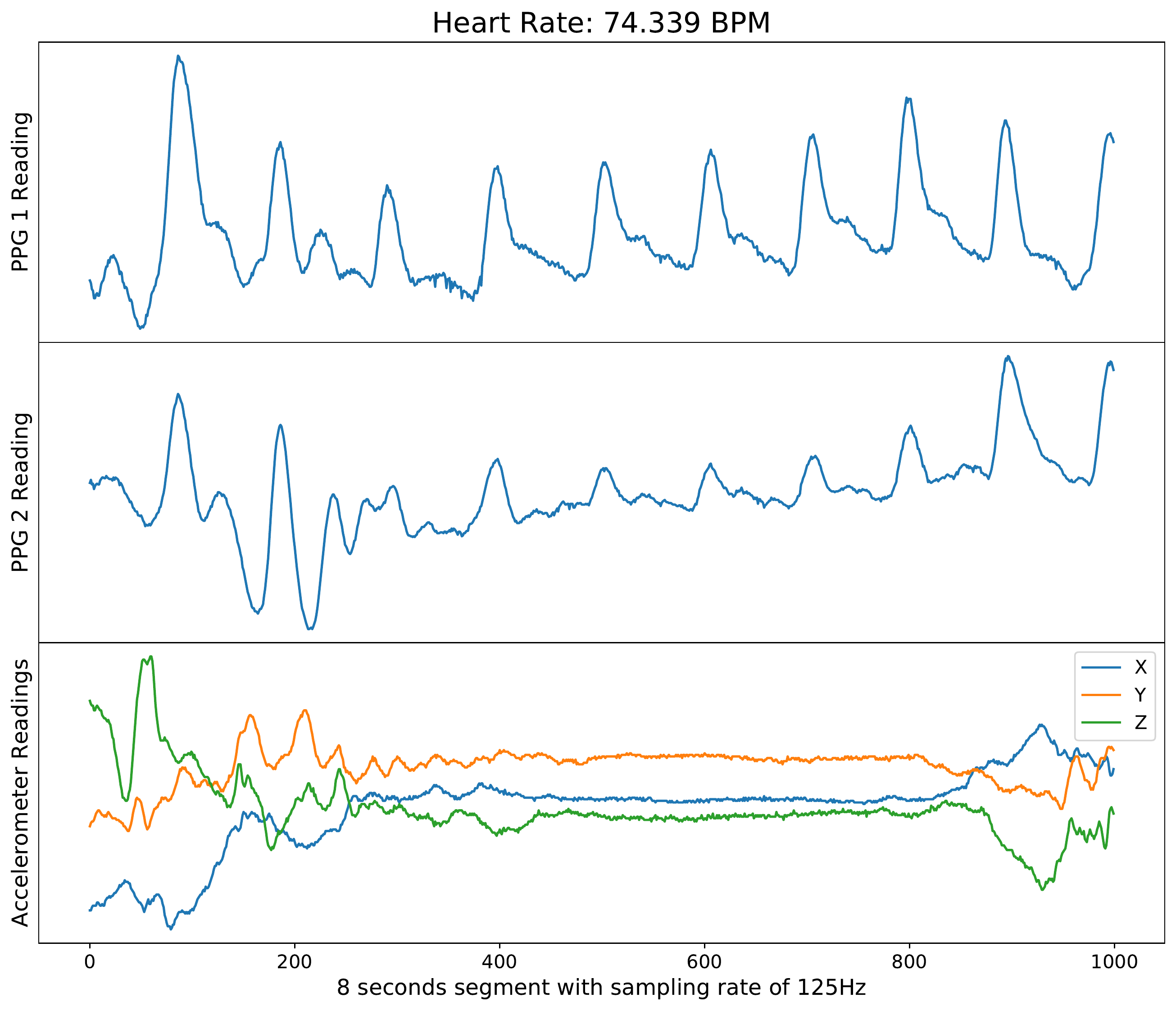}
    \caption{Example of a time series in the IEEEPPG dataset. The title shows the heart rate in beats per minute (BPM)}
    \label{fig:ieee_ppg}
\end{figure}

We modified and created the \textbf{IEEEPPG} dataset, a 5 dimensional time series from the original dataset, using the 2 PPG signals and 3-axis accelerometer signals.
The original train/test split was used, resulting in 1768 train instances and 1328 test instances. 
Similar to PPGDalia, the signals are segmented into 8 seconds windows with 6 seconds overlaps. 
With a sampling rate of 125Hz, all the time series have a length of 1000.
Figure \ref{fig:ieee_ppg} shows an example of the measurements obtained from the PPG and acceleration sensors in the dataset.

\subsubsection{The Beth Israle Deconess Medical Centre (BIDMC) PPG and Respiration}
Apart from measuring the heart rate, PPG sensors can also be used to measure other vitals such as the respiratory rate (RR) and blood oxygen saturation level (SpO$_2$) of an individual \cite{pimentel2016toward}. 
Typically PPG sensors are not very accurate in estimating respiratory rate of an individual as they fail to distinguish between periods of high and low-quality input data \cite{pimentel2016toward}.
The study by \cite{pimentel2016toward} claimed that existing systems were not robust for clinical practice.
Hence they proposed a method based on multiple autoregressive models to improve the robustness of estimating RR from PPG sensors. 
The proposed method was able to achieve comparable accuracy to existing methods whilst providing estimates for majority of the data.
They extracted a dataset\footnote{\url{https://physionet.org/content/bidmc/1.0.0/}} from the larger ``MIMIC II matched waveform Database'' that contains the physiological signals such as PPG and ECG data of the patients, sampled at 125Hz.
Then the data is manually annotated with the heart rate (HR), RR and SpO$_2$ of the patients at 1 second interval.

\begin{figure}[!h]
    \centering
    \includegraphics[width=0.7\columnwidth]{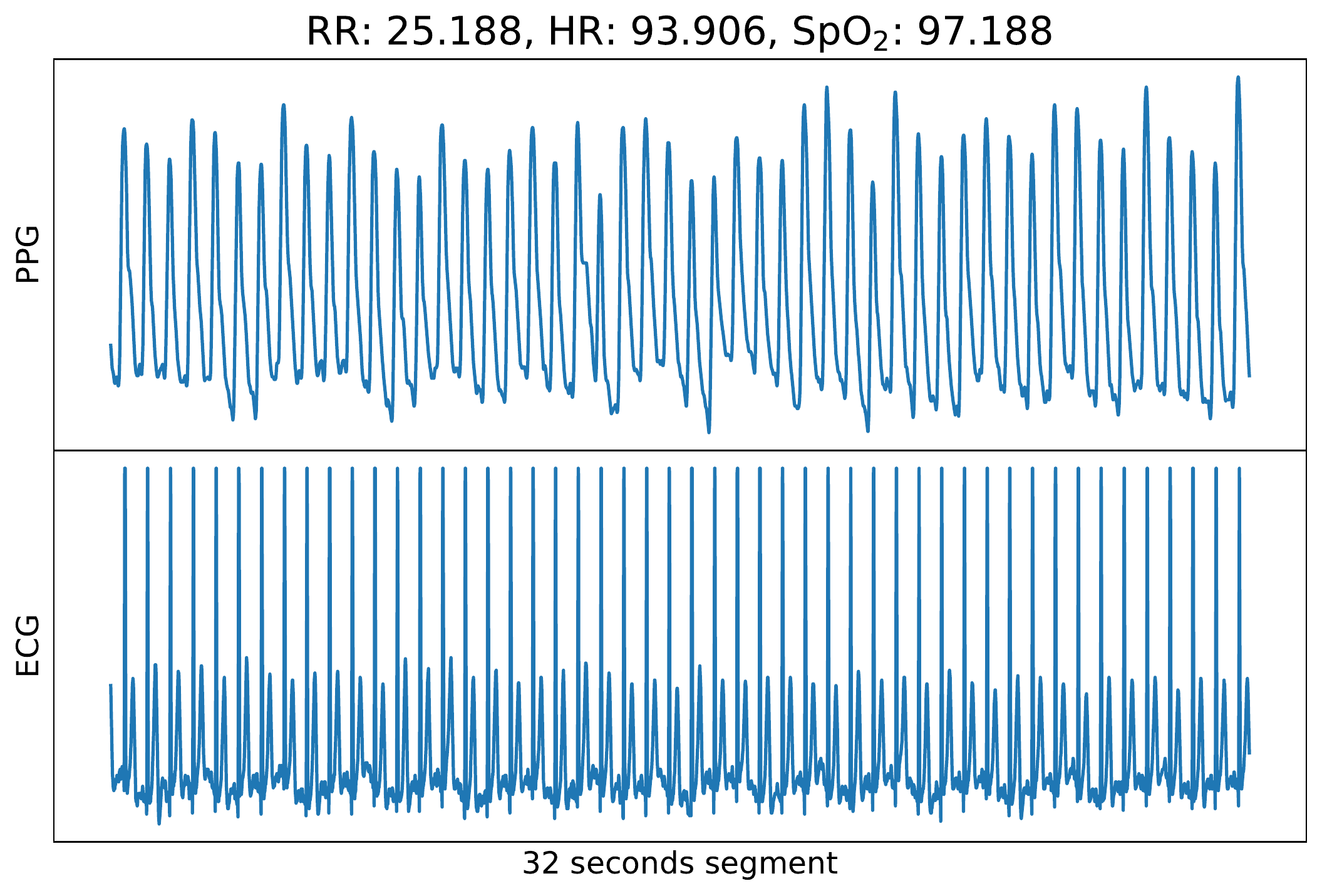}
    \caption{Example of 32 seconds PPG and ECG measurements used to estimate heart rate, respiratory rate and SpO$_2$ of a patient.}
    \label{fig:bidmc}
\end{figure}

For this archive, we adapted the original dataset from \cite{pimentel2016toward} and created the \textbf{BIDMC32HR}, \textbf{BIDMC32RR} and \textbf{BIDMC32SpO2} datasets to estimate the HR, RR and SpO$_2$ of a patient using PPG and ECG time series data.
Following the same procedure in the paper \cite{pimentel2016toward},
the PPG and ECG data were converted into time series using a 32 seconds sliding window, illustrated in Figure \ref{fig:bidmc}. 
The average HR, RR and SpO$_2$ in the 32 seconds window are used as the target for each time series. 
The datasets are split into train and test sets by randomly selecting 30\% as the test set. Therewith, BIDMC32HR consists of 5550 and 2399 train and test time series respectively; BIDMC32RR consists of 5471 and 2399 train and test time series; and BIDMC32SpO2 consists of 5550 and 2399 train and test time series.
The difference in the number of training time series is due to missing values in the annotated HR, RR and SpO$_2$ which are not included in the datasets.

\subsection{Sentiment analysis}
Sentiment analysis is the interpretation and classification of emotions (positive, negative or neutral) within some text using text analysis techniques.
This is typically done by analysing text comments or posts on websites and social media platforms \cite{moniz2018multi}.
This section describes two \emph{sentiment analysis} datasets in this archive.

\subsubsection{News popularity in multiple social media platforms}
A dataset containing 100,000 news items on four topics: \emph{economy}, \emph{microsoft}, \emph{obama} and \emph{palestine} was released by \cite{moniz2018multi} and is available in the UCI Machine Learning repository \cite{Dua:2019}.  
The dataset also contains the respective social feedback on 3 social media platforms: \emph{Facebook}, \emph{Google+} and \emph{LinkedIn}.
The dataset is collected within a period of 8 months, between November 2015 and July 2016. 
Sentiment analysis has traditionally being done using natural language processing techniques. 
Here we attempt a different approach to predict the sentiment score of news headline and news title by analysing the number of reactions on the social media platforms over a period of 2 days with time series analysis.

\begin{figure}[!h]
    \centering
    \includegraphics[width=0.75\columnwidth]{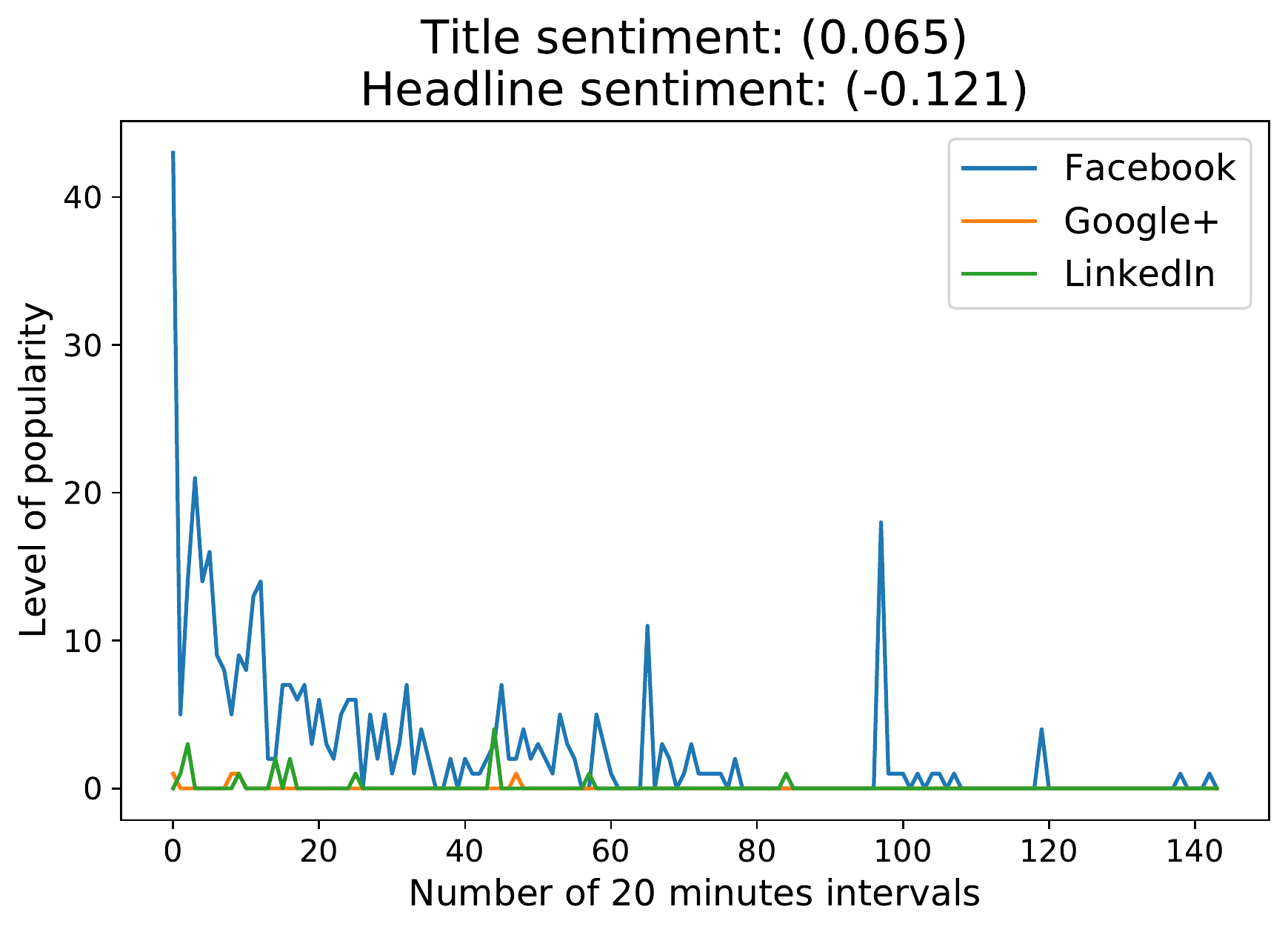}
    \caption{Example of news popularity on 3 social media platforms. The title of the news is ``Obama denounces rise of 'vulgar and divisive' politics of Trump'' with the headline ``And it's worth asking ourselves what each of us may have done to contribute to this vicious atmosphere in our politics,” Obama told the ...'' (\url{https://time.com/4259468/obama-trump-violence-rallies/}). The values in the brackets correspond to the respective sentiment value in news title and headline after 2 days.}
    \label{fig:news popularity}
\end{figure}

We created two TSER datasets \textbf{NewsHeadlineSentiment} and \textbf{NewsTitleSentiment} from the original news popularity dataset \cite{moniz2018multi}.
The datasets contain 3-dimensional time series that measure the number of reactions to the news on the 3 social media platforms.
The number of reactions was recorded at 20 minutes intervals, resulting in time series of 144 datapoints in length.
Figure \ref{fig:news popularity} shows an example of the time series in both datasets where the target variables are the sentiment scores for news headline and news title, respectively.
70\% of the dataset are randomly selected to be in the training set with 30\% in the test set, resulting in 58213 training instances and 24951 test instances. 

\subsection{Forecasting}
Time series forecasting is the task of predicting future values based on some recent and/or seasonal values.
Typically a model such as ARIMA is fitted to the historical data and extrapolated into the future.
TSER can be seen as a general case of forecasting where the goal is to predict a continuous value that may not necessarily be a future value or depending more heavily on recent values.
Thus, we included in this archive a dataset that could easily be solved with forecasting algorithms to show that forecasting tasks can also be tackled using TSER algorithms. 

\subsubsection{Covid-19}
\begin{figure}[!h]
    \centering
    \includegraphics[width=0.85\columnwidth]{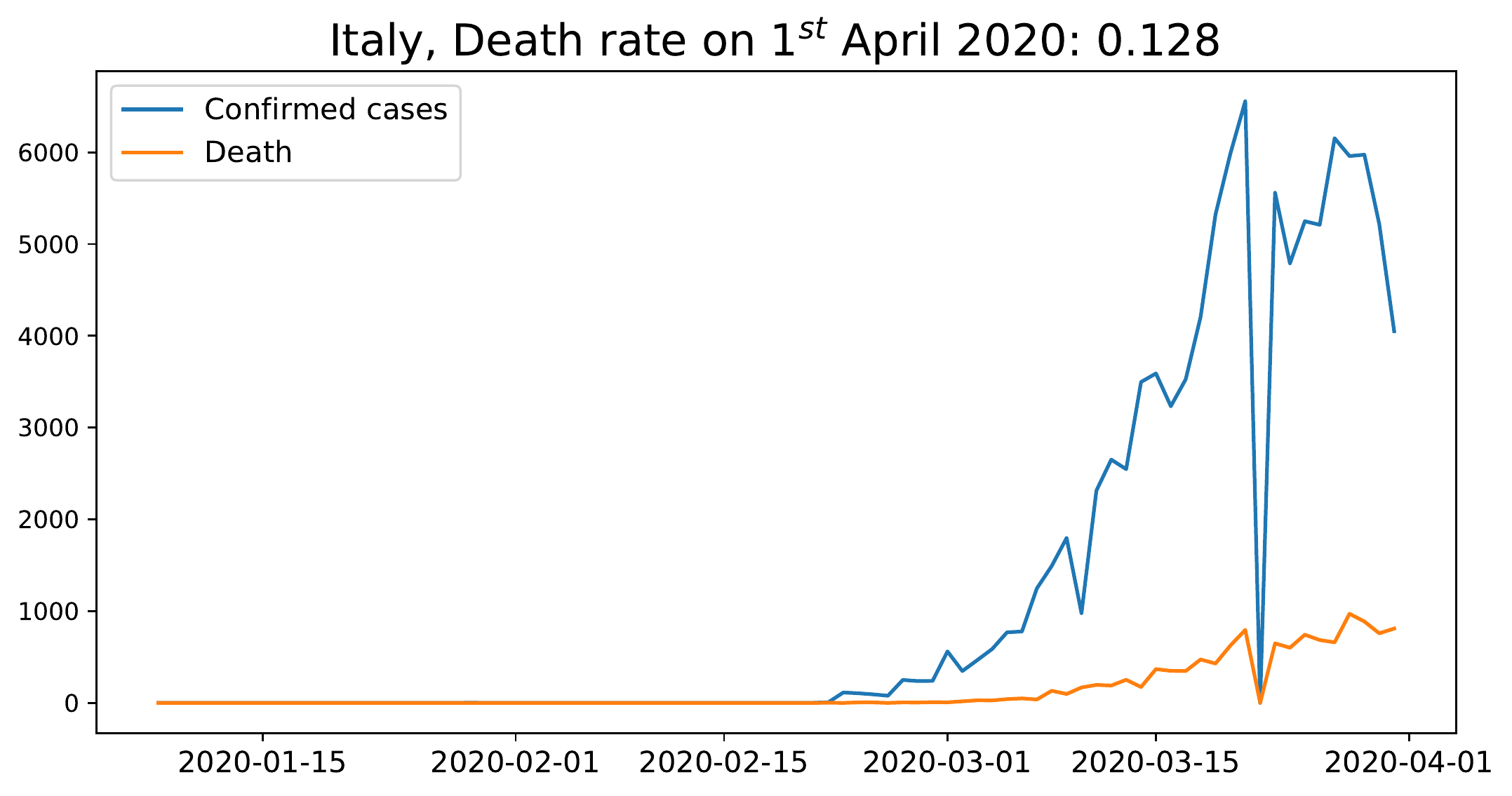}
    \caption{Daily confirmed Covid-19 cases and death rate for Italy.}
    \label{fig:covid}
\end{figure}

In 2020, the world suffers from the Covid-19 pandemic. 
Covid-19 is one of the worst pandemics in the last century. 
It is very contagious and spreads rapidly.
Within 6 months, by June 2020, there are more than 7 million cases and 430 thousand deaths worldwide.
The pandemic had also caused economy downturn for many countries.
In this archive, we created the \textbf{Covid3Month} dataset that consists of the total daily confirmed numbers of Covid-19 cases in most countries from January to March 2020.
The goal of this dataset is to predict the death rate for each country at 1 April 2020 using the daily confirmed cases for the past 3 months, illustrated in Figure \ref{fig:covid}.
Note that the death rate was terrifyingly high at 12.8\% in some countries like Italy. 
The numbers are obtained from the Covid-19 database from the World Health Organisation (WHO)\footnote{\url{https://covid19.who.int/}}. 
The dataset is split into train and test sets by randomly sampling 70\% as train. 
We note that using the number of confirmed cases is not sufficient to provide an accurate prediction and will be working on expanding the dimension of the dataset with more indicators to provide a more realistic dataset. 

\section{Baseline}
\label{sec:baseline}
In this section, we evaluate the regression algorithms described in Section 3 of our main paper \cite{tan2020time}  and set a baseline using the datasets from our TSER datasets.
We evaluate and benchmark the following regression algorithms:
\begin{enumerate}
    \item FPCR \cite{goldsmith2014estimator}
    \item FPCR with B-spline \cite{goldsmith2014estimator}
    \item Grid-search optimised SVR \cite{drucker1997support}
    \item RF \cite{breiman2001random} with 100 trees
    \item XGBoost \cite{chen2016xgboost} with 100 trees
    \item NN-ED with $k=1,5$ (1-NN-ED and 5-NN-ED)
    \item NN-DTW with $k=1,5$ (1-NN-DTW and 5-NN-DTW)
    \item FCN \cite{fawaz2019deep}
    \item ResNet \cite{fawaz2019deep}
    \item Inception Network \cite{fawaz2020inceptiontime}
    \item Rocket \cite{dempster2020rocket}
\end{enumerate}
Missing values in the time series are linearly interpolated. 
When using a traditional regression algorithm (i.e. non-temporal), the time series are flattened out into a single long feature vector.

We used the standard Scikit-Learn Python library \cite{scikit-learn} to implement SVR and RF algorithms. 
The SVR algorithm is optimised by performing a 3-fold cross validation with grid search on the hyper-parameters.
Specifically, the kernel, gamma and $C$ parameters are optimised from a standard range of values. 
The kernel function is selected from RBF and Sigmoid. 
The gamma parameter selected from $[0.001, 0.01, 0.1, 1]$ defines the influence of support vectors. 
The regularisation parameter $C$ is selected from $[0.1, 1, 10, 100]$.
XGBoost was implemented using the Python XGBoost library\footnote{\url{https://xgboost.readthedocs.io}}.
Apart from the number of trees, we use the default parameters for both RF and XGBoost from the Python libraries. 
Our empirical experiments show that RF and XGBoost with parameters optimised using a grid search strategy performs similarly or worse than the default parameters and takes a significantly longer time to train. 
Hence they are excluded from this benchmark.
The FPCR and FPCR with B-spline models are implemented using the Scikit-FDA Python package\footnote{\url{https://fda.readthedocs.io}}, a library for functional data analysis in Python.

For time series algorithms, we adapted the code from \cite{fawaz2019deep}\footnote{\url{https://github.com/hfawaz/dl-4-tsc}} for both ResNet and FCN and \cite{fawaz2020inceptiontime}\footnote{\url{https://github.com/hfawaz/InceptionTime}} for Inception Network.
The code for Rocket was taken from \cite{dempster2020rocket}\footnote{\url{https://github.com/angus924/rocket}} and modified for multivariate time series with the help from the original authors. 
The multivariate version of Rocket applies the transformation to each dimension independently.

The time series nearest neighbours algorithms were all implemented in Java. 
Our source code has been made open source online at \url{https://github.com/ChangWeiTan/TSRegression}. 

Since some of the algorithms are non-deterministic, we evaluate all the algorithms over 5 runs and report the average root mean squared error (RMSE), one of the most widely used metrics for regression tasks. 
Equation \ref{eqn:rmse} describes the formal definition of RMSE where $n$ is the number of instances, $y_i$ and $\hat{y_i}$ are the actual and predicted target respectively.

\begin{equation}
    RMSE = \sqrt{\frac{\sum_{i=1}^n (\hat{y_i}-y_i)^2}{n}}
    \label{eqn:rmse}
\end{equation}

We compare the algorithms statistically over the current datasets following the recommendations from \cite{demvsar2006statistical}.
First, we rank each algorithm by RMSE for every dataset.
Rank 1 is assigned to the algorithm with the lowest RMSE while rank 13 is assigned to the highest one.
Fractional ranking is assigned to the algorithm in case of ties.
We then compute the average rank for each algorithm.
Then, the Friedman test \cite{friedman1940comparison,demvsar2006statistical} was applied to the average ranks.
If the null hypothesis is rejected, the post-hoc two-tailed Nemenyi test is used to compare the algorithms to each other \cite{demvsar2006statistical}.
Using this test, the performance of the algorithms is significantly different if the average ranks differ by at least the critical difference shown in Equation \ref{eqn:cd}, where $q_{\alpha}=3.313$ is the critical value for $\alpha=0.05$, $k=13$ being the number of algorithms and $N=19$ being the number of datasets.
This gives $CD=4.186$.

\begin{equation}
    CD = q_{\alpha}\sqrt{\frac{k(k+1)}{6N}}
    \label{eqn:cd}
\end{equation}



\begin{figure}[!h]
    \centering
    \includegraphics[width=\columnwidth]{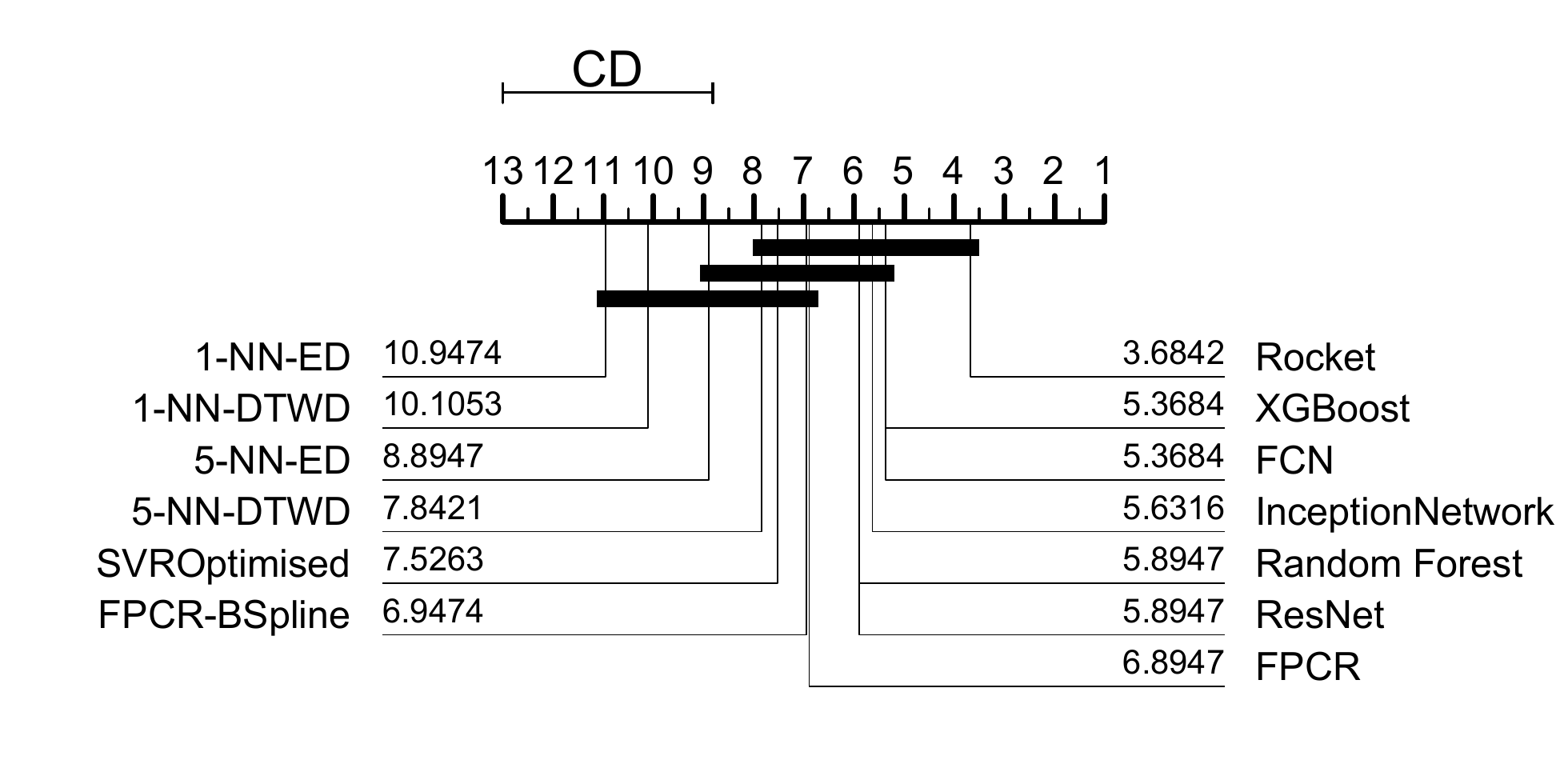}
    \caption{Critical difference diagram showing statistical difference comparison of regression algorithms on the current regression archive}
    \label{fig:cd}
\end{figure}

Finally, a critical difference diagram was used to visualise the comparison, where the thick horizontal line connecting a group of algorithms indicates that all the algorithms in the group are not significantly different from one another \cite{demvsar2006statistical}. 
Figure \ref{fig:cd} shows the critical difference diagram of comparing the algorithms used to benchmark the existing archive.
The average ranks are indicated next to the algorithms in the figure. 

Figure \ref{fig:cd} shows that Rocket is the most accurate algorithm with an average rank of 3.6842 and is significantly different to SVR, NN-ED and 1-NN-DTWD.
The figure also shows that there is no significant difference between the state-of-the-art time series algorithms and the classical regression algorithms.
This suggests that a better algorithm needs to be developed for TSER problems.
We refer interested readers to our paper \cite{tan2020time} for a more detailed discussion of the results.

\section{Conclusion and Future Work}
\label{sec:conclusion}
We have released the first iteration of the TSER archive that contains 19 time series datasets, and set an initial baseline on the archive using typical machine learning regression and state-of-the-art TSC algorithms. 
Our results show that Rocket, one of the state-of-the-art TSC algorithms performs the best overall.
State-of-the-art machine learning algorithms such as XGBoost and Random Forest are very competitive as well.
This suggests that better algorithms need to be developed for such TSER problems.
Finally, we welcome any donations of data and will continue to expand the archive, providing a wider range of problems. 

\section*{Acknowledgement}
This material is based upon work supported by the Air Force Office of Scientific Research, Asian Office of Aerospace Research and Development (AOARD) under award number FA2386–18–1–4030.
The authors would like to thank the authors of \cite{fawaz2019deep} and \cite{dempster2020rocket} for providing their source code online. 
The authors also appreciate the data donation from all the donors.

\bibliographystyle{ieeetr}
\bibliography{references}

\end{document}